\documentclass[graybox]{svmult}

\usepackage{helvet}         
\usepackage{courier}        
\usepackage{type1cm}        
\usepackage{makeidx}         
\usepackage{graphicx}        
\usepackage{multicol}        
\usepackage[bottom]{footmisc}

\makeindex             

\begin{document}

\title*{It is Time for New Perspectives on How to Fight Bloat in GP}
\author{Francisco Fern\'andez de Vega  \and
       Gustavo Olague \and
       Francisco Ch\'avez \and
       Daniel Lanza \and
       Wolfgang Banzhaf \and
       Erik Goodman
}
\institute{F. Fern\'andez de Vega, F. Ch\'avez, D. Lanza \at University of Extremadura, Spain \email{fcofdez, fchavez\{@unex.es\}, dlanza1@gmail.com}
\and G. Olague \at Cicese, M\'exico \email{olague@cicese.mx}
\and W. Banzhaf, E. Goodman \at Beacon Center, MSU, Usa \email{banzhafw,goodman\{@msu.edu}\}}
%
%

\authorrunning{Fern\'andez et al.}

\maketitle

\abstract*{The present and future of evolutionary algorithms depends on the proper use of modern parallel and distributed computing infrastructures.  Although still sequential approaches dominate the landscape, available multi-core, many-core and distributed systems will make users and researchers to more frequently deploy parallel version of the algorithms.  In such a scenario, new possibilities arise regarding the time saved when parallel evaluation of individuals are performed.  And this time saving is particularly relevant in Genetic Programming.   This paper studies how evaluation time influences not only time to solution in parallel/distributed systems, but may also affect size evolution of individuals in the population, and eventually will reduce the bloat phenomenon GP features.
This paper considers time and space as two sides of a single coin when devising a more natural method for fighting bloat.  This new perspective allows us to understand that new methods for bloat control can be derived, and the first of such a method is described and tested. Experimental data confirms the strength of the approach:  using computing time as a measure of individuals' complexity allows to control the growth in size of genetic programming individuals.}

\abstract{The present and future of evolutionary algorithms depends on the proper use of modern parallel and distributed computing infrastructures.  Although still sequential approaches dominate the landscape, available multi-core, many-core and distributed systems will make users and researchers to more frequently deploy parallel version of the algorithms.  In such a scenario, new possibilities arise regarding the time saved when parallel evaluation of individuals are performed.  And this time saving is particularly relevant in Genetic Programming.   This paper studies how evaluation time influences not only time to solution in parallel/distributed systems, but may also affect size evolution of individuals in the population, and eventually will reduce the bloat phenomenon GP features.
This paper considers time and space as two sides of a single coin when devising a more natural method for fighting bloat.  This new perspective allows us to understand that new methods for bloat control can be derived, and the first of such a method is described and tested. Experimental data confirms the strength of the approach:  using computing time as a measure of individuals' complexity allows to control the growth in size of genetic programming individuals.}

\section{Introduction}
\label{intro}
A well known phenomenon in GP is the inherent bloating behavior correlated with fitness improvement  \cite{fitnessandbloat}.  Many approaches to fix this problems have been described and applied although no perfect solution has yet been found.

This paper considers the problem from a new point of view.  The main goal is to understand whether new paths are possible, so that in the future new bloat control methods can be produced.  Instead of using more traditional size-based approaches (such as penalty functions associated to chromosome size), we are particularly interested in analyzing whether the influence of parallel and distributed computing models that reduce computing time may be also useful for reducing bloat.  Although island models have  been analyzed before in this context (\cite{fernandez1}, \cite{whigham}), these previous approaches relied more on spatial structure of the models, while we are here more interested in the standard GP algorithm, when it is run in parallel using the standard fitness parallelization approach.

The analysis that we present is useful to understand the relationship between individual size and computing time, and therefore with the bloat phenomenon.  Moreover, a new set of bloat-control mechanisms could be easily derived, that makes use of parallel architectures that are nowadays present in every computer system.

We thus analyze the bloat phenomenon using execution time instead of memory consumption (size).  To the best of our knowledge this is the first time such approach is considered and applied.  As we describe below, preliminary tests with a well known benchmark problem shows the feasibility of the idea.

Thus, the main contribution of this chapter is providing a new perspective for bloat fighting in GP, and any variable-size chromosome based evolutionary approach;  secondly, we describe how this perspective may inspire new bloat-control methods for GP; and finally, one of such control methods is described and tested with success.  Although results are still preliminary, we are confident with results obtained, which will be confirmed in the future with a series of experiments in a wider set of benchmark problems.

The rest of the chapter is organized as follows:  In Section \ref{bloat} we contextualize the problem, and then, their relationship with parallel architectures and scheduling are described in Section \ref{load}.  The methodology applied is presented in Section \ref{methods}, while Section \ref{Results} shows the experiments performed and results obtained.  Finally, we draw our conclusions in Section \ref{Conclusions}.

\section{The Bloat phenomenon}\label{Bloat}
\label{bloat}
The bloat problem has been addressed frequently in the GP literature since first described by J. Koza in  \cite{koza1992}.   A good review of the topic may be found in  \cite{bloatreview}.  We will refer here to ideas that have been described in the last decade and are more directly related with the approach we follow in this chapter.



Among the available techniques to control bloat, particularly relevant is a recent method called \textit{the waiting room},  introduced by \cite{waitingroom}.  The idea is for individuals to add a pre-birth phase to all newly created individuals.  Children must wait for a period of time proportional to their size before they are allowed to enter the population and compete.  Although the authors recognized that the idea was associated with the relationship between individual sizes and evaluation times, they maintained the emphasis on size-control mechanism and hence did not elaborate on the time concept, nor did they take into account the possibilities associated with parallel and distributed infrastructures available, given their influence on evaluation time when individuals are sent to available processors.  Thus, they relied on the total number of nodes individuals feature, similarly to all of the other methods, although using a somewhat different approach.  

Another technique of interest is  \textit{operator equalization} presented in \cite{dignum} aimed at controlling the distribution of program sizes at each generation, defining a specific shape for the distribution.  Some of the best results were achieved by using a uniform or flat distribution (\cite{silva}), and also by applying speciation, fitness sharing or elitism, see \cite{trujillo}. But again, difficulties in effectively applying the method include how to control the shape of the distribution without changing the nature of the search, and how to efficiently account for individuals' sizes and shapes.

Although plenty of size-related techniques can be found in the literature for bloat control in GP, few times, if any, a computing time analysis have been tackled.   Here we come up with a deeper analysis of computing times that sheds light on the problem, in contrast to the standard approach of individuals' sizes.  Moreover, we want to see if the relationship between size and evaluation time can be exploited in parallel systems not just to save time, but to address the bloat phenomenon in a more natural way.

We must also remember that in a series of papers, it was observed that the island model offers some possibilities for fighting bloat (\cite{fernandez1}, \cite{fernandez2}, \cite{fernandez3}), and this observation was later exploited in a new proposal by Whigham \cite{whigham} that considers spatial distribution of islands in GP. The connection between the dynamics of some of the parallel models for GP and the bloat phenomenon has been shown to be mainly due to the spatial structure of the model, which relies on islands of individuals.  But there is still a second source of possible improvement in parallel EAs, as we described above: the number of computing resources employed to run the algorithm, which has not been studied yet from the point of view of its influence on algorithm's bloating behavior.  Even when the simplest embarrassingly parallel model is used for running a GP experiment, a load balancing technique must decide how individuals are distributed among the available computing resources, and this may also have an influence on the bloat phenomenon, as we show below.

\section{Load-Balancing and parallel GP}\label{Load}
\label{load}
Among the available parallel models for EAs and GP, the only one that does not change the algorithm behavior is the embarrassingly parallel model. All algorithm's steps are performed as in the sequential version, and the only change is introduced in the most expensive part of the algorithm:  the fitness evaluation.  Thus, instead of sequentially evaluating every individual in the population, they are distributed among the available processors, and fitness values are computed in parallel.

This model is frequently used in the parallel computing literature, being known as the client/server model.  It requires some kind of load-balancing mechanism that allows to reduce latencies and distribute tasks efficiently among computing resources, so that \textit{makespan} is reduced.  Interested readers can find a taxonomy of load-balancing methods in \cite{Osman}, while \cite{Zaki} presents a comparison of different strategies.

If we focus on GP, given that a large number of individuals featuring different sizes must be managed across the available processors, which are typically smaller in number than the population size, some kind of load-balancing mechanism must be applied, in charge of sending individuals to idle processors, and this mechanism might provide new hidden properties:  sometimes, a deeper analysis of the new version of a given algorithm allows us to discover some properties that were not noticed before. We are here interested in both the parallel model itself, and the load-balancing technique that can be used and considered as the basis for a new proposal that we will describe and analyze below.

Load-balancing techniques have already been considered as an implicit component of parallel versions of genetic programming.  Since the nineties, static load-balancing mechanisms --the ones we will consider here-- have been applied within parallel versions of GP, when facing difficult real-world problems. For instance, \cite{oussaidene} describes a parallel version of GP that considers complexity of individuals as the basis for establishing the load-balancing policy. 

Nevertheless, few papers since then have studied the importance of load-balancing techniques in GP. We may refer to \cite{preliminaryloadbalancing}, where several methods were tested. But again, no specific study on their relationship with the bloat phenomenon has already been described. 

\subsection{Structural complexity of GP individuals}

When any load-balancing technique is to be employed, a prediction of computing time for the task must be applied, so that the method can properly decide when to launch the task.  In GP, an important feature of GP individuals is their structural complexity (\cite{complexity}).  Although this value is typically computed taking into account the number of nodes, such as the case of evaluating \textit{computing effort} (\cite{phd}) or lexicographic complexity (\cite{lexicographic}), both are approximate estimates of the real value required: in other words, the evaluation time of the individuals' fitness function.

But we can adopt a different point of view, as we do in the approach we present below: given that the estimation of the real complexity of an individual is measured when the individual has been evaluated, we can characterize individuals using that computing time, so that we can employ it in future decisions.  In any case, that value will not be available when the load-balancing mechanism must decide when to launch the evaluation of a new individual; nevertheless, it will be available after the individual's evaluation. This could be useful, if not for that individual, given that it was already sent to be evaluated, then at least for its children, as a value to somehow approximate its evaluation time. 

And this is basically the idea we will apply:  Our approach takes into account an individual's computing time, as a value to decide how to distribute children among available computing resources, and ultimately, to reduce computing time while simultaneously reducing the bloat phenomenon.   We thus need to keep a record of the time that each program spends during testing and use that information to create clusters of programs with similar durations that will be useful for load-balancing individuals:  clusters will be send to different processors.  Thus, the load-balancing mechanism relies on individuals' computing time, and this allows to return individuals after evaluation in an order that depends on their computing time.  We use the computer's clock to give a value to runtime of a program; this, of course, is correlated with the size of the computer program and the number of instruction cycles required to execute it. All clusters are created without regard to the fitness function. We do not measure directly the size of the individuals, nor use any information about the complexity of breeding programs other than time.   

As we show below, the above described load-balancing technique has an impact on the bloat phenomenon without increasing the computational complexity of the algorithm.

\section{Methodology}\label{Methods}
\label{methods}

As described above, we will consider execution time as a measure of an individual's complexity given that a correlation exists between size and running time, that can be more deeply investigated in future work.


This idea can be easily applied when individuals are evaluated: we just have to take elapsed time during an individual's evaluation as the complexity value required. 


The idea is particularly useful when multicore or manycore computer architectures are to be employed:  ideally, all of the individuals in the population could be evaluated simultaneously, and their evaluation time obtained simultaneously.

We simplify the measurements by directly using the elapsed evaluation time as the representation of the individual's complexity.

\begin{figure}[!t]
\centering
\includegraphics[width=3.8in]{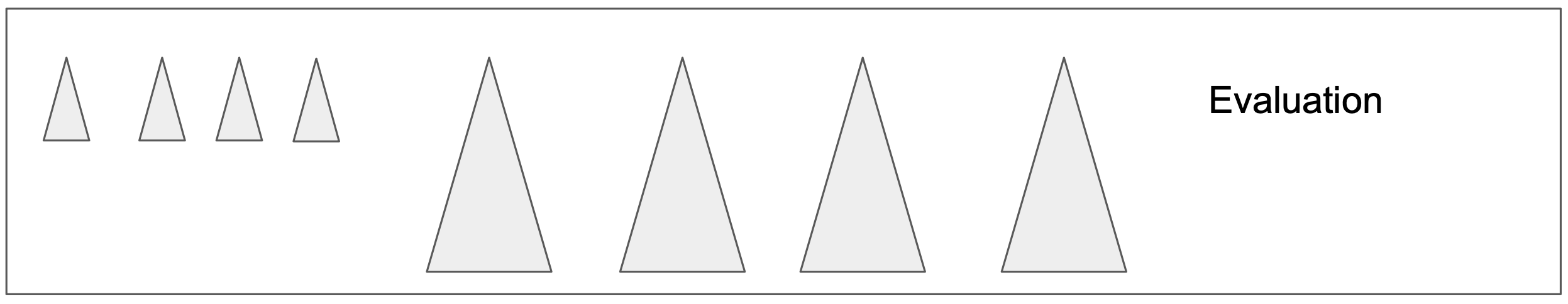}
\caption{First step:  Individuals are sent to be evaluated.}
\label{step_1}
\centering
\includegraphics[width=3.8in]{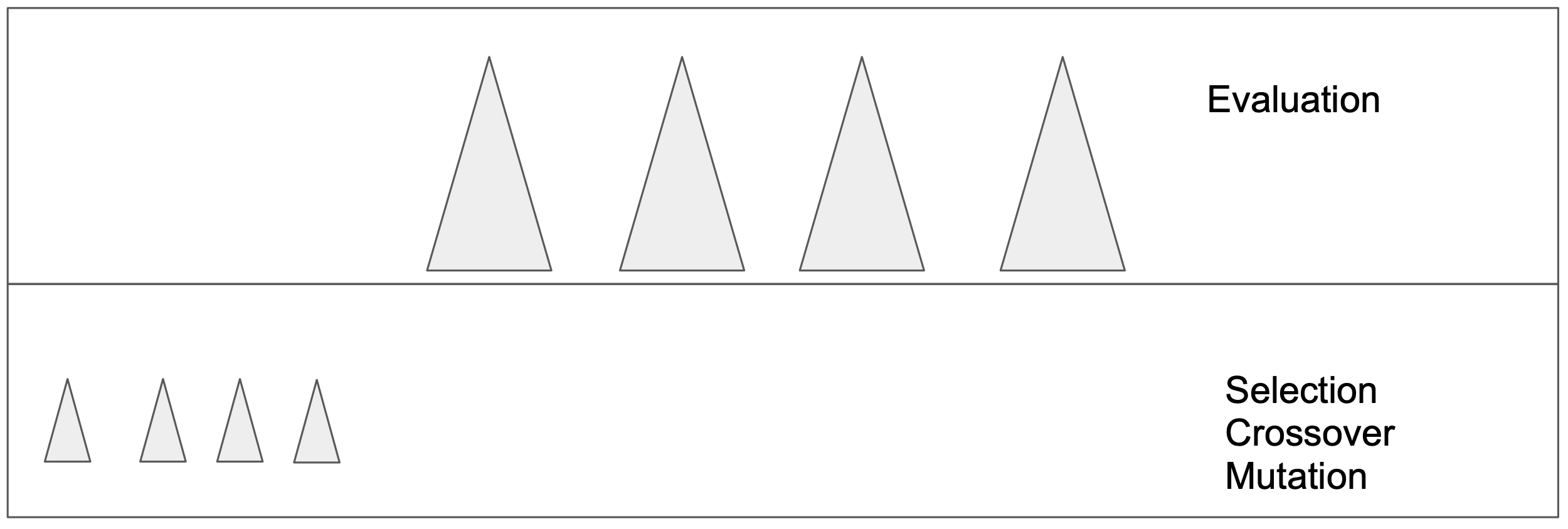}
\caption{Second step:  Smaller individuals fitness available first.}
\label{step_2}
\centering
\includegraphics[width=3.8in]{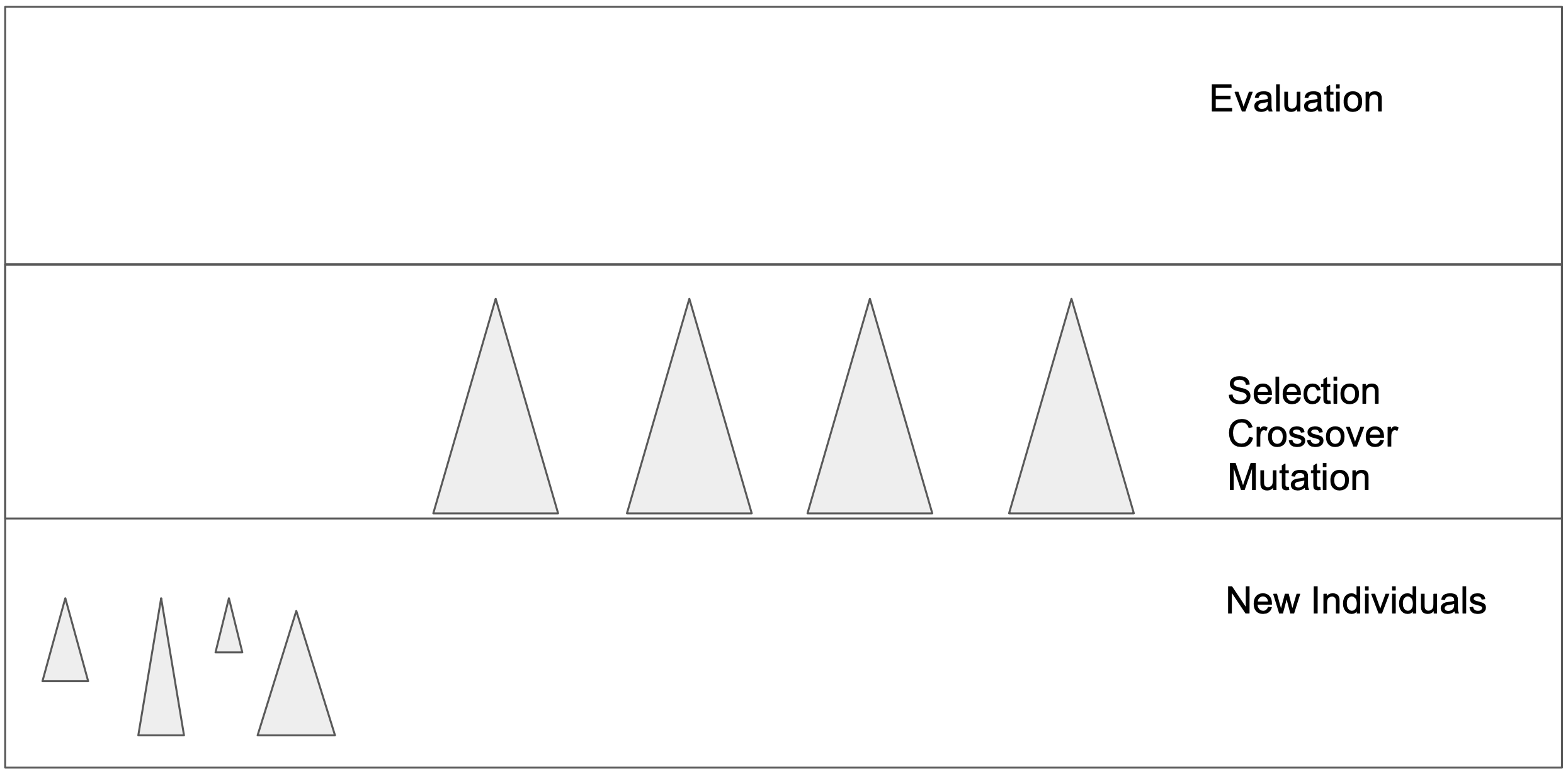}
\caption{Third step:  Smaller individuals produce children while larger ones are coming back from evaluation.}
\label{step_3}
\centering
\includegraphics[width=3.8in]{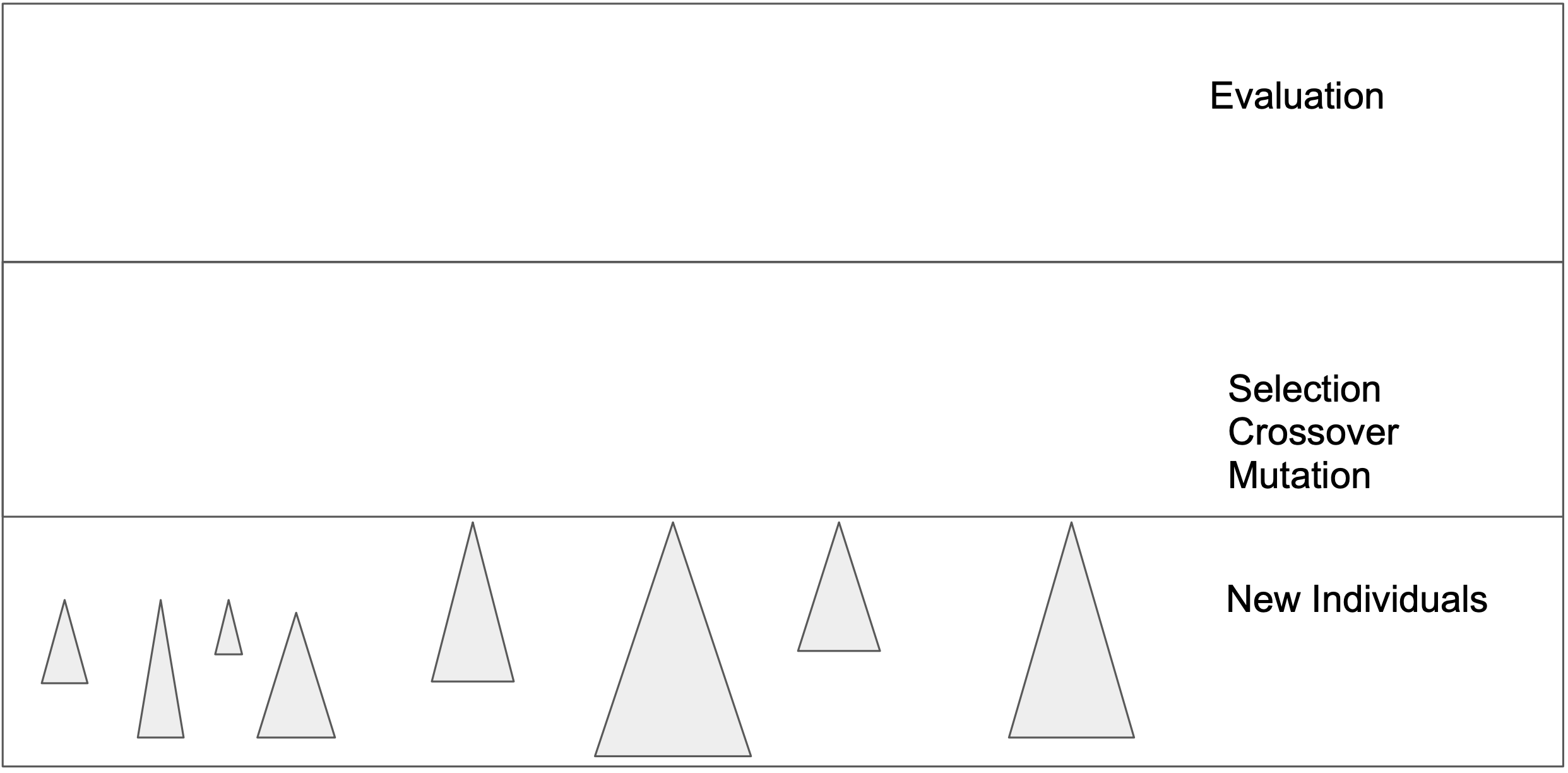}
\caption{Fourth step:  Larger individuals produce children.}
\label{step_4}
\end{figure}

Once the individuals' evaluation times have been obtained, and with the hypothesis that individuals of similar size will produce offspring of similar size, our proposed method groups individuals by computing time, always understanding it as an indirect --and easier to compute-- measure of an individual's size.  We must again consider that in a parallel system with as many processors as individuals, individuals of similar size will finish their evaluation simultaneously and will be ready to reproduce.  Therefore, an automatic grouping mechanism naturally arise from these parallel architectures (see figures \ref{step_1}-\ref{step_4}).  If the number of processors is smaller, then, the load balancing mechanism which is always in charge of distributing tasks among processors, will decide which individuals group together in single tasks, and may thus apply grouping according to the ending time of individuals evaluation.

After grouping, selection and breeding phases are performed within each group, so only individuals of similar size-time are allowed to crossover. Then, the load-balancing mechanism is in charge of creating tasks by grouping individuals of the same cardinality by evenly dividing the whole population.

Our hypothesis states that individuals of similar size will produce offspring of similar size. This will not be the case if the crossover operation does not divide the individual into two similar-size parts. If different, crossover will produce small and big individuals, whose sizes do not follow our hypothesis. Nevertheless, considering that individuals are randomly divided, we expect a central tendency which results aim to be of a size in between both parents. This can be considered as a weak point of our offspring-size prediction, which can be improved in a future version. However, generally speaking, we expect that offspring will have similar sizes compared to their parents.

\begin{figure}[!t]
\centering
\includegraphics[width=3.8in]{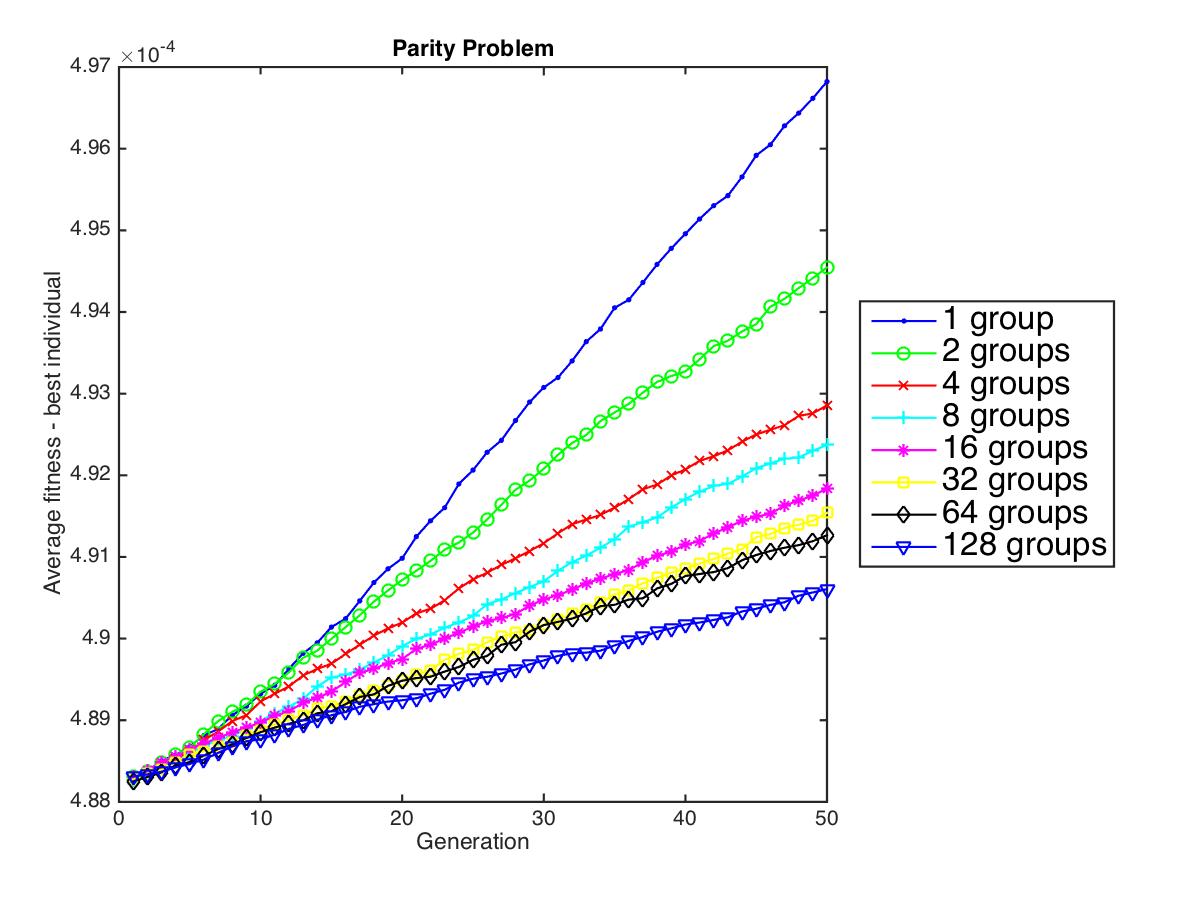}
\caption{Best-fitness evolution along generations (averaged over 30 runs) for the parity problem (maximizing fitness).}
\label{parity_avg_fitness}
\end{figure}

\subsection{Implementation}


Therefore, when designing a specific bloat control mechanism for GP that uses the new time-based perspective, the parallel version of the algorithms, in particular multi-thread based models, has been initially chosen, which are available today in some of the most popular EAs tools.  Thus, all operations performed within each group are done in one thread, and the number of threads created corresponds to the number of groups. Each thread collects its corresponding individuals and performs the selection and breeding steps. In this way, all operations are isolated within each group, that can be naturally conformed when individuals return from evaluation according to evaluation time.

After the breeding phase, the mechanism takes advantage of the fact that each group of individuals is contained in a different thread and it performs the evaluation of all the corresponding individuals. Each group/thread contains the same amount of individuals, individuals of similar execution time. As a result, the evolutionary process is considerably speeded up by parallelizing the evaluation phase. Afterwards, once all individuals of the population are evaluated, and their computing times obtained, they are sent to the thread corresponding to their computing time (size surrogate) value, so that the next breeding operations can be performed. 


\subsubsection{Software tool}

With the goal of making the bloat method easily usable, it has been implemented based on a popular existing tool, ECJ \cite{ecj}. Such system has been built in modules in order to facilitate the replacement of any part involved in the evolutionary process. In our case, we replace the module that carry out the breeding phase with the new time-based approach.

Our bloat control mechanism slightly modifies the way individuals are bred. 
In order to apply the bloat control mechanism, two 
new operations have been implemented. 

\begin{itemize}
  \item \textit{GroupBreeder} orchestrates the breeding phase and starts the corresponding threads. 
    \begin{itemize}
      \item As the first step, individuals need to be grouped according to evaluation times. During their evaluations, elapsed time has been captured so each individual already contains it as a new feature. Before grouping them, all the individuals of the population are sorted by evaluation time.
      \item Then, the same number of individuals goes to each group, so they are taken in order and sent to their groups. In case individuals cannot be equally split into groups, first groups will get one individual more.
      \item Next, one thread is instantiated per group. A group of individuals is assigned to each of them.
      \item Threads are started and the program continues until all threads have finished.
    \end{itemize}
  \item \textit{GroupBreederThread} represents the threads that actually perform the selection, breeding and evaluation of individuals. 
    \begin{itemize}
      \item A call to the module that performs the selection and breeding is done, specifying the group to which these operations need to be applied. Here, the only change that has been done to the original implementation is to apply these operations to only the individuals that correspond to the specified group -- the group that corresponds to the thread.
      \item Once selection and breeding phases have finished, evaluation of new individuals generated by this thread takes place.
    \end{itemize}
\end{itemize}

Note that the evaluation 
step has not been modified. Therefore, it goes through all individuals and tries to evaluate them; however, it will not actually evaluate any, since they have been previously evaluated and marked as such.

\subsection{Experiments}




All experiments described below were run on an Intel(R) Xeon(R) CPU (E5530) that offers 8 cores at 2.4 GHz and 8 GB of memory. Default configuration parameters set in ECJ has been used for the benchmark problem. Generations were set to 50, and 30 runs were launched, for statistical purposes. The well known even parity problem from the GP literature was used with the basic configurations already available in the ECJ toolkit. The only change is the number of bits in the chromosome:  12 bits so that the problem is difficult enough for long runs.  Regarding the load-balancing mechanism -number of groups to be used- several configurations were employed: 1 group, which corresponds with the standard GP algorithm, and also 2, 4, 8, 16, 32, 64, and 128 groups.




\section{Results}\label{Results}
\label{results}
We present and discuss below results obtained in the experiments.  Average fitness and size are plotted on the figures included. The proposed method uses parallel execution and results are presented here: as many threads as possible are launched so that individuals of different sizes are evaluated in different processors;  Yet, ideas extracted can be also adapted when running experiments in a sequential fashion.  We also include an analysis of results in this context.  

\subsection{Parallel model}

\begin{figure}[!t]
\centering
\includegraphics[width=3.8in]{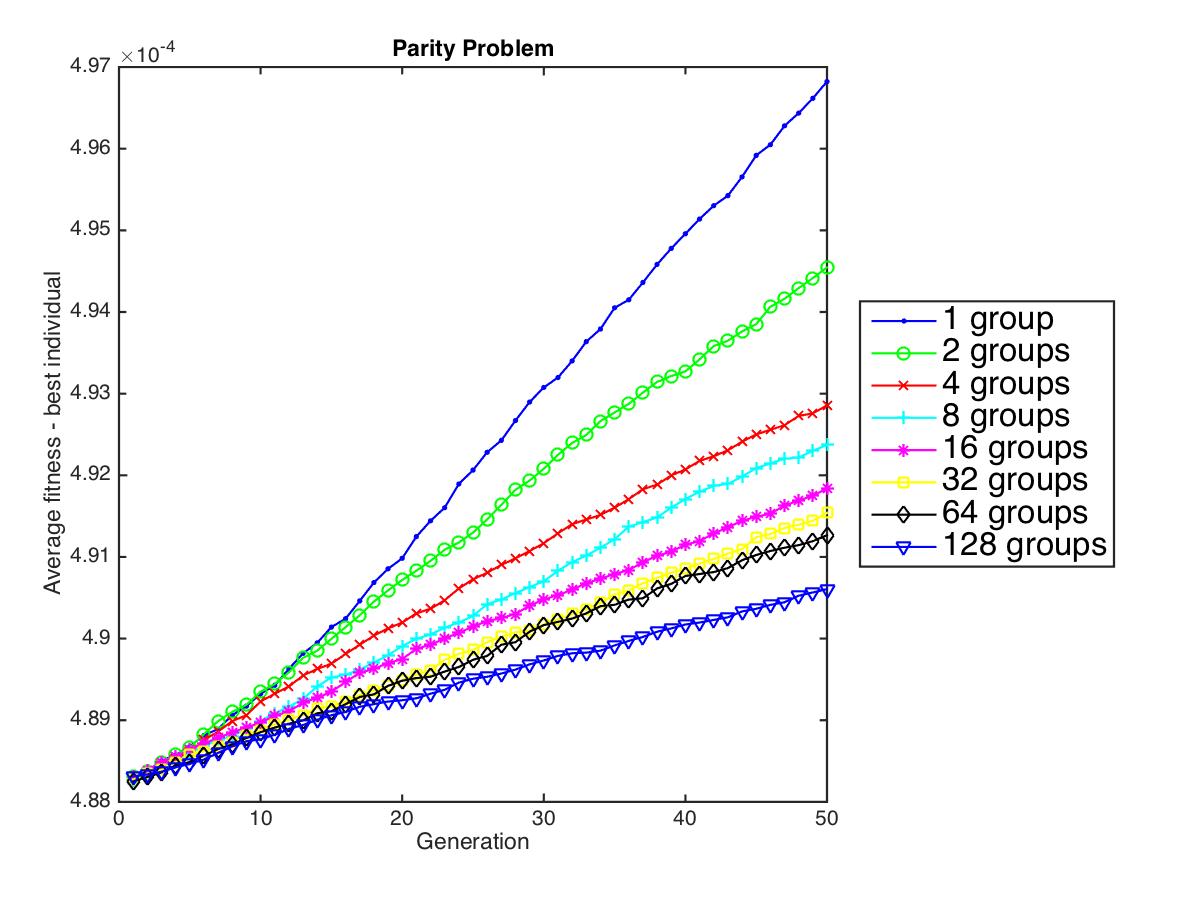}
\caption{Best-fitness evolution along generations (averaged over 30 runs) for the parity problem (maximizing fitness).}
\label{parity_avg_fitness2}
\end{figure}

In the parity problem, fitness is monotonically affected by the number of threads (groups), as can be observed in Figure \ref{parity_avg_fitness2}. Nevertheless, if we take into account the scale employed, differences are really narrow.

\begin{figure}[!t]
\centering
\includegraphics[width=3.8in]{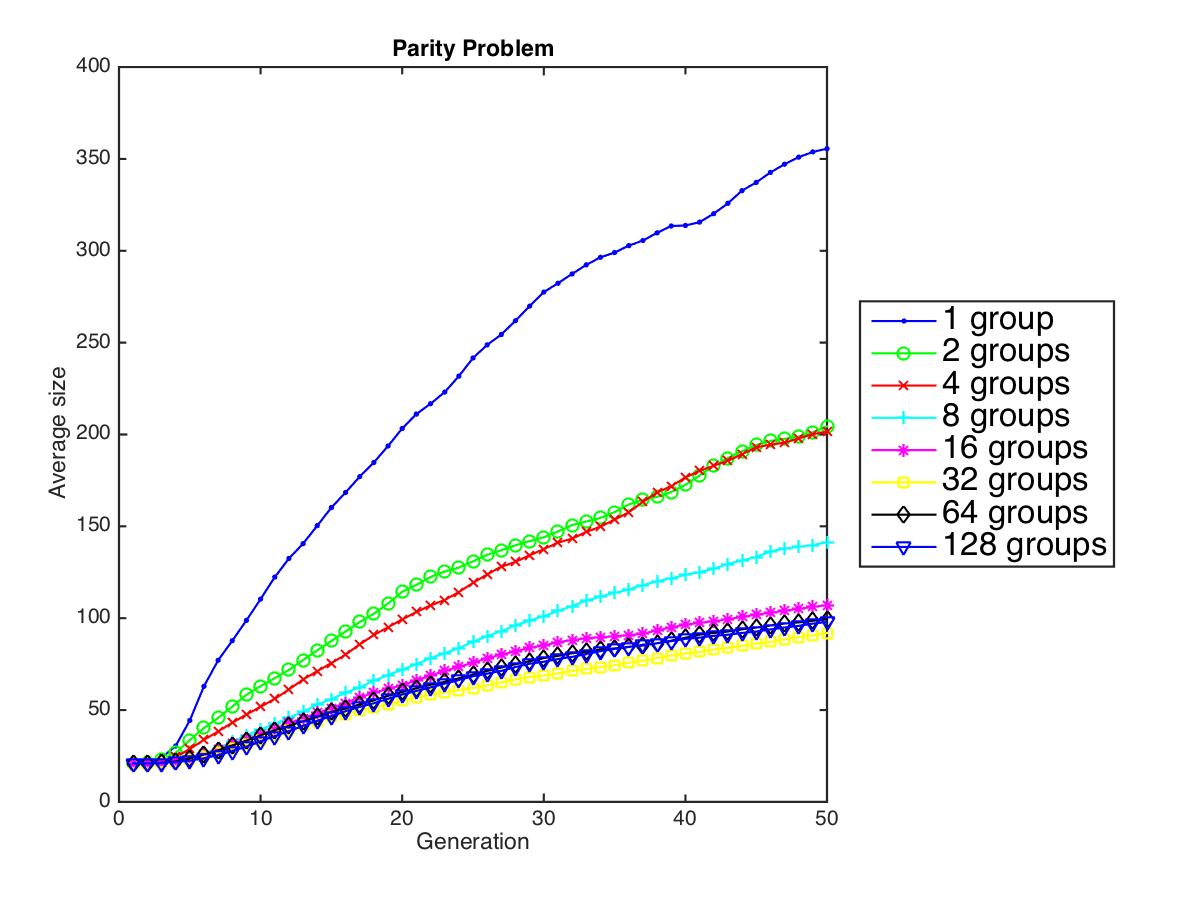}
\caption{Size-evolution along generations (averaged over 30 runs) for the parity problem.}
\label{parity_avg_size}
\end{figure}

If we focus instead on size evolution, a dramatic reduction up to a third of the size as compared with a standard run (1 group) can be found in Figure \ref{parity_avg_size}. The slightly affected fitness may be acceptable, taking into account the considerable reduction in size.

\subsection{Sequential execution}

Although the previously described idea was born considering how individuals can be run on parallel computing systems, given that individuals may naturally group when a number of them are launched to be evaluated on different processors, and return simultaneously when their running time is similar, the idea can be adapted to sequential environments with minimal changes: allowing individuals to group according to running time.  Unfortunately delays are present in this latest approach, given that individuals can only be grouped when all of them has been evaluated, while in the parallel model this is not the case:  they can breed with other individuals that has finish their evaluation process simultaneously.  In any case, the idea for the sequential version is to simply emulate the parallel version.  Therefore, no population structure support the model, although some resemblances with structured models may be seen in this sequential approach.  But this distinction is pertinent to properly understand the new bloat control approach:  while the method naturally fits parallel computing environments, and can be applied without any additional effort, other structured-based approaches can only be applied when specific grouping tasks are added to the algorithm.



\begin{figure}[!t]
\centering
\includegraphics[width=3.8in]{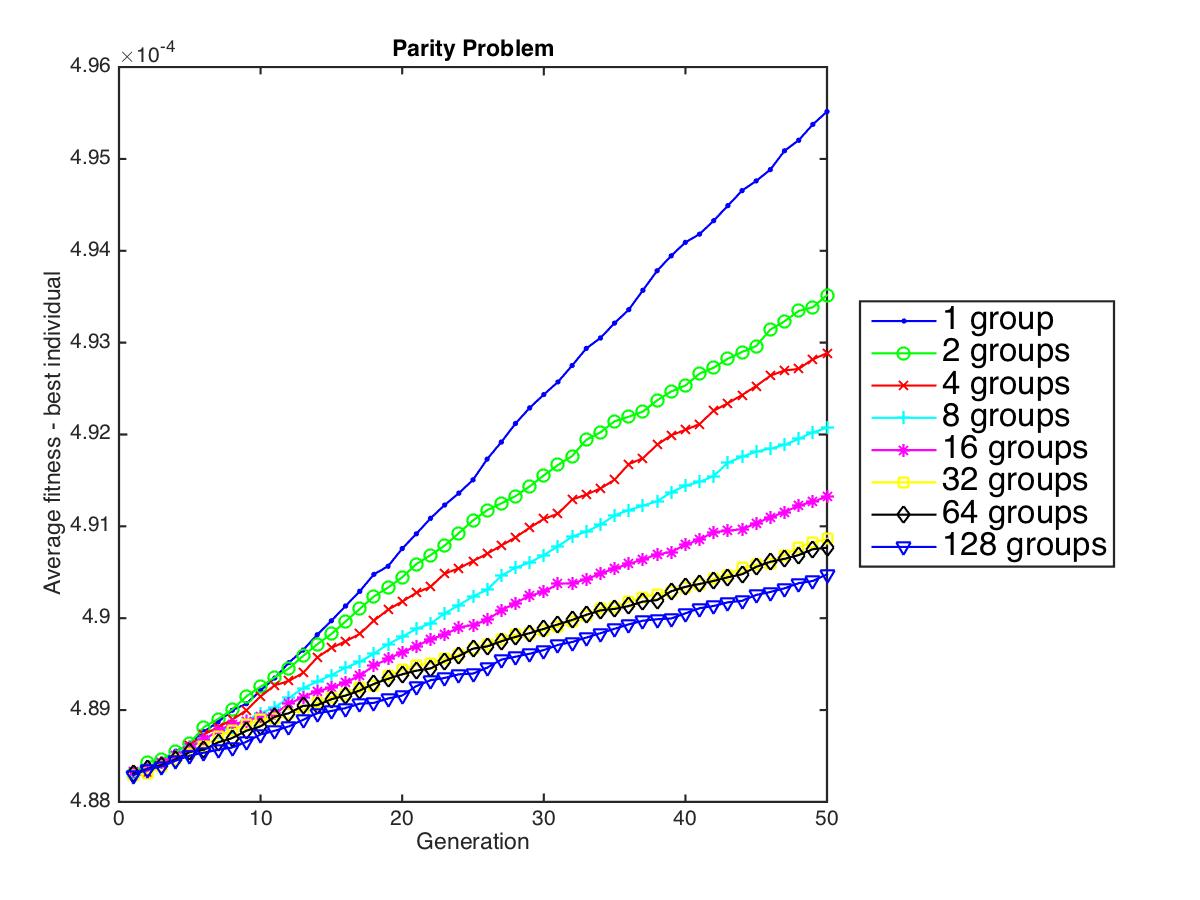}
\caption{Best-fitness evolution along generations (averaged over 30 runs) for the parity problem (sequential execution) (maximizing fitness).}
\label{parity_avg_fitness_sequential}
\end{figure}

\begin{figure}[!t]
\centering
\includegraphics[width=3.8in]{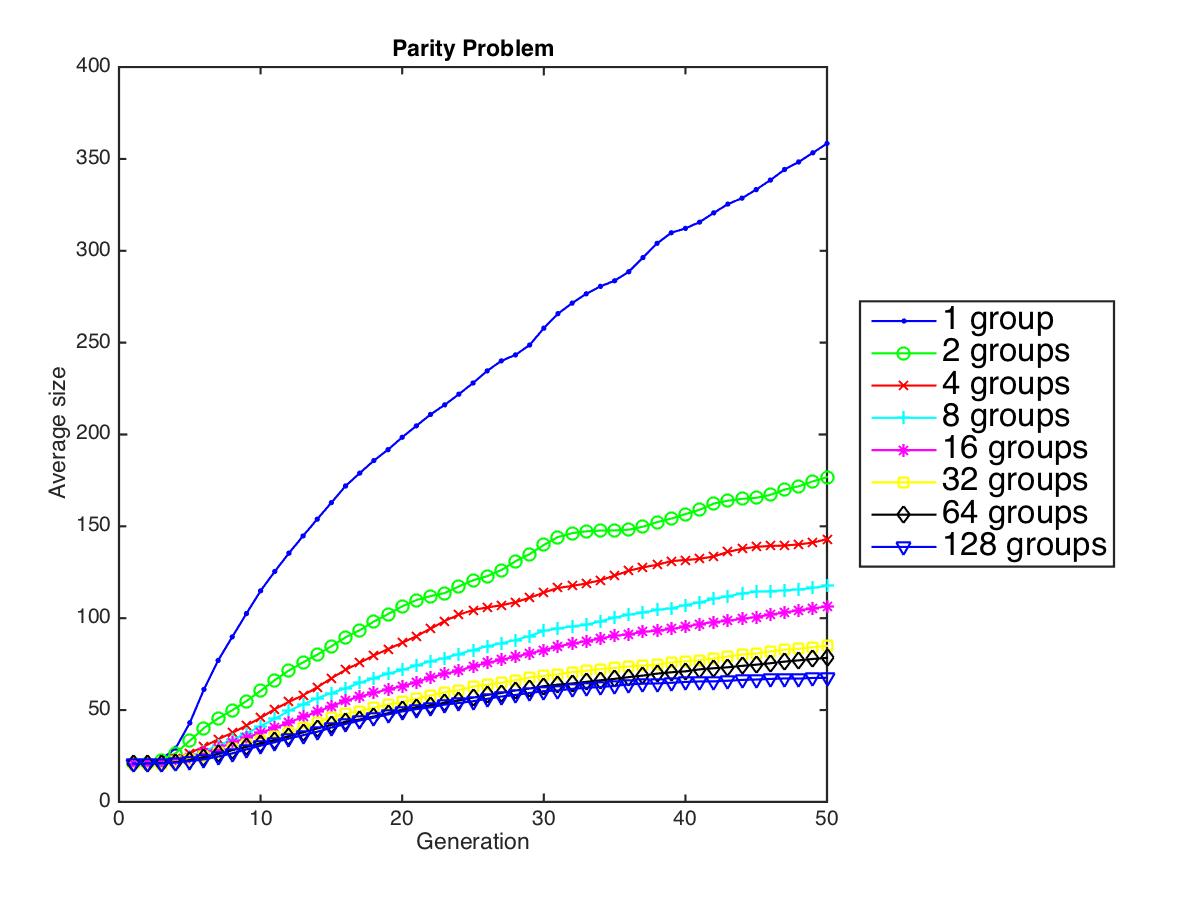}
\caption{Size-evolution along generations (averaged over 30 runs) for the parity problem (sequential execution).}
\label{parity_avg_size_sequential}
\end{figure}

Results from sequentially executed runs are shown for the even parity problems in figure \ref{parity_avg_fitness_sequential} and \ref{parity_avg_size_sequential}. Similarly to what is seen in the results from parallel execution, fitness quality is not strongly affected, while the individual size is notably reduced. This phenomenon is produced solely by the fact of grouping individuals by computing time before carrying out selection and crossover stages.


Summarizing, we have shown in this preliminary study that the 
proposed time and individual duration method points to new research avenues where GP could be studied to address the problem of size growth from a computing time perspective. Moreover, under the new approach this paper introduces a first method for controlling individual sizes. Its main idea was to group individuals according to evaluation time, and it naturally allows keeping individuals' size under control, while fitness quality remains high. The experiments allow us to see the interest of the approach in both sequential and parallel models, although it requires further analysis with a larger set of problems in future work to confirm the findings presented above.

Although we have focused here on the time-size relationship as well as on the specific method implemented under this perspective, we believe that new approaches may be developed in the future considering time-space relationships in variable-size chromosome-based evolutionary techniques.  Moreover, we think that specific improvements in the method presented here for GP will be attained when parallel environments are used and a proper load-balancing approach is applied.

\section{Conclusions}\label{Conclusions}
This chapter presents a new approach to evaluating individuas complexity in variable-size chromosomes based evolutionary algorithms:  using computing time instead of individuals size.  Moreover, given that in parallel/distributed computing environment load-balancing methods may allow individuals to naturally group according to arrival time, this idea can shed light on the bloating phenomenon, providing clues for new control methods.





To demonstrate the usefulness of the approach, we present a first method which is based on an individual's computing time --which is automatically obtained when fitness is computed-- as a trait employed for characterizing and grouping individuals together in a natural way, so that they can only bred within their groups. The reason for this idea is to keep computing time --and thus, indirectly, an individual's size growth-- under control. 

Based on the above described idea, we have run a set of experiments on a well known benchmark problem:  parity, and results -both in parallel and sequential environments- show that the idea works, and a first specific method to prevent bloat has been presented, although other possible ones that relies in load-balancing techniques may be derived.



%
\begin{acknowledgement}
We acknowledge support from Spanish Ministry of Economy and Competitiveness under project TIN2017-85727-C4-f2,4g-P, Regional Government of Extremadura, Department of Commerce
and Economy, the European Regional Development Fund, a way to build Europe, under the project IB16035, Junta de Extremadura, project GR15068, and CICESE project 634-128.
\end{acknowledgement}
\bibliographystyle{spmpsci}      
\bibliography{ecjolague}   
\end{document}